# An Integrated Forecasting Prototype for Emergency Department Boarding Time to Support Proactive Operational Decision Making

Orhun Vural[1], Abdulaziz Ahmed[2,3], Ferhat Zengul[2], James Booth[4], Bunyamin Ozaydin[2,3, *]

[1] Department of Electrical and Computer Engineering, University of Alabama at Birmingham, Birmingham, AL, USA
[2]Department of Health Services Administration, University of Alabama at Birmingham, Birmingham, AL, USA
[3]Department of Biomedical Informatics and Data Science, Heersink School of Medicine, University of Alabama at Birmingham, Birmingham, AL, USA
[4]Department of Emergency Medicine, University of Alabama at Birmingham, Birmingham, AL, USA

## Abstract

Overcrowding in emergency departments (ED) remains a persistent operational challenge worldwide, causing delays in care delivery and downstream congestion. ED boarding time, defined as the duration admitted patients remain in the ED while awaiting inpatient bed placement, is a key indicator of this congestion. Predicting ED boarding time in advance enables proactive operational decision making before congestion escalates. We developed and evaluated a multi-horizon time series forecasting framework to predict ED boarding time at 6, 8, 10, 12, and 24-hour horizons. Real-world data from a university-affiliated urban hospital in the United States were utilized and integrated with external contextual data sources, including weather, holidays, and major local events. Decomposition-based Linear (DLinear) and Normalization-based Linear (NLinear) time series forecasting deep learning models showed superior performance across multiple horizons. Models were also evaluated under extreme congestion scenarios characterized by elevated boarding times. In addition, a Machine Learning Operations (MLOps) web application prototype was developed to support translation of the forecasting framework into practice through integrated data ingestion, forecast visualization, experimentation, and retraining.

## Introduction

Emergency Department (ED) overcrowding is an ongoing and globally recognized challenge associated with delays in care, reduced quality, increased costs, adverse outcomes, operational strain, and ambulance diversion when capacity is exceeded [1, 2]. Operationally, ED crowding reflects a system-level imbalance in patient flow, where demand surpasses available capacity, resulting in prolonged waiting, delayed treatment, and extended boarding due to inpatient bed constraints [3]. To quantify congestion, prior research focuses on patient flow metrics (PFMs), which are occasionally referred to as key performance indicators (KPIs) in the literature. These include time-based measures (e.g., length of stay, door-to-doctor time, waiting time, boarding time, diversion time), count-based indicators (e.g., waiting, treatment, and boarding counts), and resource utilization and productivity metrics (e.g., bed occupancy, staff utilization, leave without being seen), among other related operational indicators reported in the literature [4, 5].

Boarding time is a commonly used indicator of ED overcrowding, reflecting both delays in timely care and constraints in inpatient capacity [4, 6]. Boarding time is defined at the encounter level as the interval beginning when an inpatient bed request is placed by the treating physician and ending when the patient checks out of the ED for transfer to an inpatient unit [7]. Admitted patients who wait in the ED for inpatient beds occupy treatment rooms and reduce the department's ability to care for incoming patients, and this prolonged boarding has been associated with higher mortality and morbidity [8, 9]. Consequently, boarding time serves as a critical exit-side bottleneck in ED patient flow, with cascading effects that extend from patient arrival to final disposition. In a recent U.S. hospital setting, the average daily cost per medical–surgical patient boarding in the ED was USD 1,856, compared with USD 993 for comparable inpatient care, underscoring the substantial financial burden of prolonged ED boarding [10].

A common strategy to mitigate ED overcrowding is the Full Capacity Protocol (FCP), a hospital-wide surge plan that increases throughput through overflow activation, accelerated discharge, and coordinated staffing [11]. However, FCP is typically activated only after overcrowding occurs, making it reactive rather than proactive. Predictive, data-driven approaches have therefore emerged to anticipate congestion in advance and enable earlier operational adjustments [12]. Most models treat ED patient flow as a time-series process, using temporally ordered PFMs to forecast future conditions and guide resource allocation [13]. External contextual factors, such as temperature, holidays, air quality, flu activity, and major events are often incorporated to capture fluctuations in demand [14, 15]. Recent advances also leverage unstructured Electronic Health Record (EHR) data, including triage notes and clinical narratives, alongside patient-level demographic, socioeconomic, and clinical variables [16]. Despite diverse feature sets, no standardized predictors exist, as optimal inputs vary across hospitals and operational contexts [17]. Although interventions such as FCP are operational rather than software-based, translating them from reactive to proactive practice requires more than accurate prediction alone. It also requires an integrated application environment that can automate data extraction, generate forecasts in a timely manner, present outputs to operational users, and support ongoing model maintenance and retraining.

ED overcrowding has been examined from multiple perspectives, ranging from high level patient flow concepts to detailed operational performance metrics. More recently, the field has shifted toward data driven forecasting, where machine learning methodologies are increasingly deployed to predict key PFMs such as arrival count [18, 19], waiting time [20], inpatient admission [21], boarding volume [22], and length of stay [23, 24], to support proactive operational decision making. A recent scoping review confirmed that machine learning–based models consistently outperform traditional statistical methods, such as rolling-average approaches and linear regression [25-27].

Prior research has examined ED boarding mainly through statistical, interventional, and simulation-based approaches, with limited use of predictive machine learning. Regression-based studies identified associations between longer boarding times and patient complexity, insurance status, and departmental congestion, as well as downstream increases in inpatient length of stay [28, 29]. Interrupted time series and simulation frameworks have evaluated admission protocols and discharge timing strategies to mitigate boarding [30, 31]. More recently, count-based representations of boarding have been used as system-level congestion signals within forecasting models, including deep learning time-series approaches for hourly boarding counts and machine learning models for predicting extended boarding events [14, 32]. Overall, prior studies on ED boarding have primarily relied on descriptive analyses, regression modeling, interventional evaluations, or simulation-based approaches. Although machine learning approaches have been applied to related congestion indicators such as boarding count and aggregated patient flow metrics, we did not identify prior studies that directly model and forecast ED boarding time as a continuous outcome. In addition, prior work has given limited attention to how such forecasting can be translated into an integrated operational application that supports automated data processing, timely prediction, and ongoing model maintenance.

This study developed a predictive framework to forecast ED boarding time across short and medium range horizons (6, 8, 10, 12, and 24 hours ahead) using data from a university affiliated urban hospital to support proactive operational planning. Boarding time was aggregated hourly by summing patient boarding minutes and averaging across boarded patients present during each hour. Benchmarking of state-of-the-art time-series models was conducted to assess performance across short- and medium-range horizons. Models were built using routinely available operational and system-level flow data, without patient-level clinical variables, and incorporated engineered hourly features to capture overall ED state. Performance was further evaluated during statistically defined of extreme congestion periods. Finally, we developed a Machine Learning Operations (MLOps)-enabled web prototype to operationalize the forecasting framework within a unified application environment. The prototype was designed to automate and streamline key steps required for practical use, including data streaming, feature generation, forecast visualization, experimentation, and model retraining, thereby supporting feasibility assessment and early user feedback prior to deployment. In this way, the study addresses not only the predictive challenge of forecasting ED boarding time, but also the translational challenge of embedding such forecasts within an application capable of supporting more proactive operational responses.

# Results

## Data characteristics

Real-world internal operational data from a university-affiliated urban partner hospital in the United States, including ED tracking data source and inpatient data source, were integrated with external contextual data from weather services [33], federal holiday calendars [34], and football event schedules [35, 36] to capture ED flow dynamics and external demand influences. These same data sources were also used in our prior forecasting studies of ED patient flow metrics, including waiting count and boarding count [14, 37]. All data sources cover the period from January 1, 2019 to July 1, 2023.

As detailed in the Data Source and Feature Engineering subsection of the Methodology, 29 features were constructed for model development. The feature set included five temporal features, consisting of year, month, day of the month, day of the week, and hour. From the ED tracking dataset, fourteen features were constructed, including boarding time, boarding count, ESI specific boarding counts for levels 1 and 2, level 3, and levels 4 and 5, waiting time, waiting count, ESI specific waiting counts for levels 1 and 2, level 3, and levels 4 and 5, treatment time, treatment count, total patient count, and an extreme boarding indicator. Two inpatient features were included: census count and surgical count. Weather features consisted of temperature and categorical indicators for five weather conditions, including clouds, rain, thunderstorm, clear, and other. In addition, two event related features were incorporated to represent football game days for two local teams. The dataset was structured at an hourly resolution, with each row corresponding to one hour, resulting in a total of 37,191 observations. Descriptive statistics for all features are presented in Figure 1, including mean, standard deviation (SD), minimum, and maximum values, with percentages reported for categorical variables.

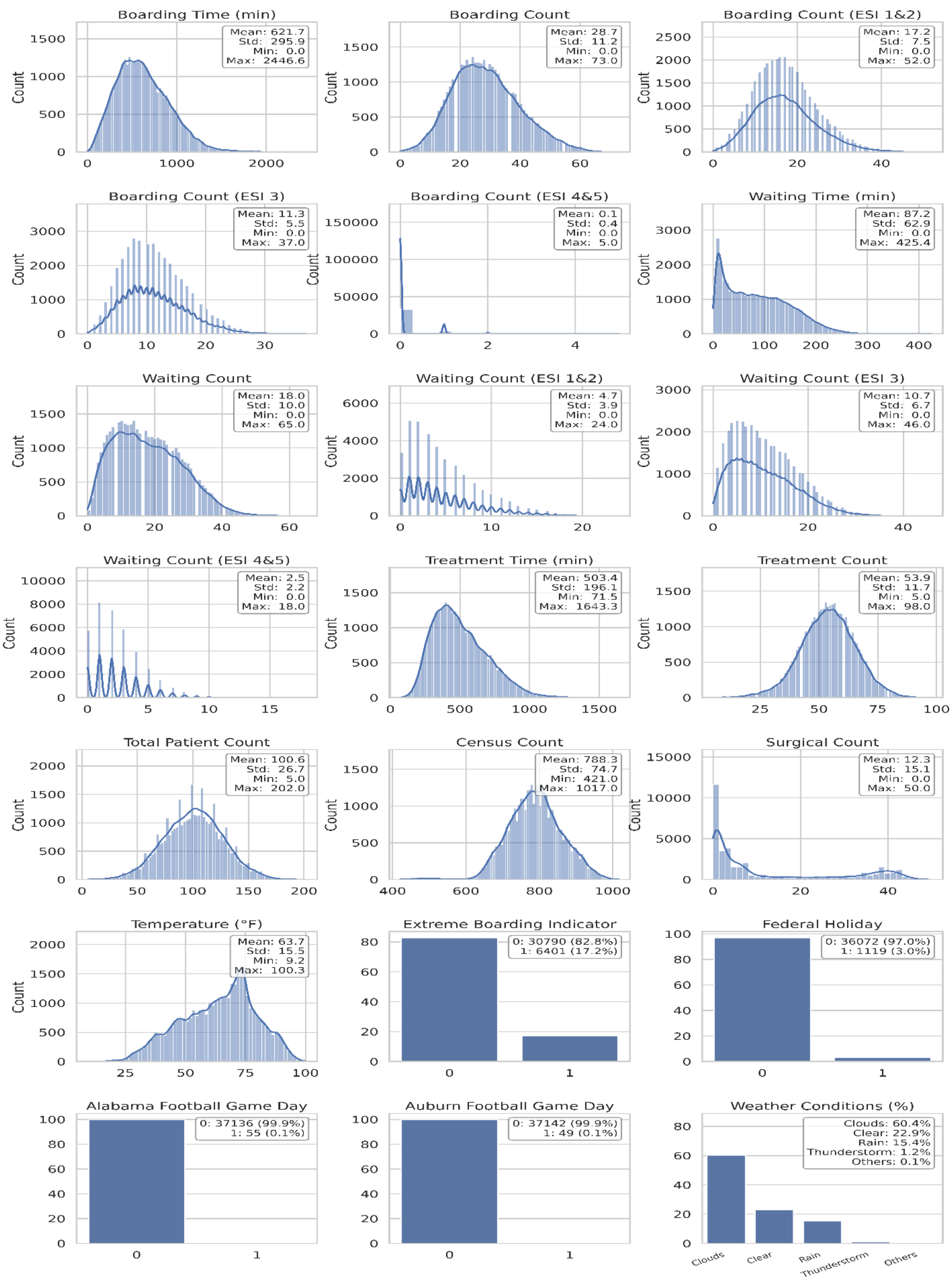

**Fig. 1 |** Distributions and summary statistics of features utilized in this study, including ED operational and external contextual indicators, at an hourly resolution.

## Model Performance

A comprehensive evaluation of predictive performance was conducted for boarding time forecasting models. This included assessing predictive performance at 6, 8, 10, 12, and 24 hour forecast horizons and benchmarking five time series architectures: Time Series Dense Encoder (TiDE) [38], Decomposition Linear (DLinear) [39], Normalization Linear (NLinear) [39], Temporal Fusion Transformer (TFT) [40], and Time Series Transformer Plus (TSTPlus) [41]. Forecasting accuracy was evaluated using mean absolute error (MAE), root mean square error (RMSE), coefficient of determination ($R^2$), and mean absolute percentage error (MAPE) to capture both absolute and relative prediction performance.

As shown in Figure 1, the mean boarding time across the study dataset was 622 minutes, with a standard deviation of 296 minutes. Figure 2 presents the comparative forecasting performance of all evaluated algorithms across the different prediction horizons. Overall, NLinear achieved the best performance across the shorter forecast horizons. Specifically, at the 6-hour horizon, NLinear obtained the lowest MAE (94.2), with an RMSE of 122.6, an $R^2$ value of 0.8, and a MAPE of 13.3%. At the 8-hour horizon, NLinear again produced the lowest MAE (105.0), accompanied by an RMSE of 135.7, an $R^2$ of 0.7, and a MAPE of 14.9%. This pattern continued at the 10- and 12-hour horizons, where NLinear recorded MAE values of 115.4 and 122.9, RMSE values of 148.2 and 158.3, $R^2$ values of 0.7 and 0.7, and MAPE values of 16.3% and 17.5%, respectively. At the longest forecast horizon of 24 hours, DLinear achieved the strongest performance, with the lowest MAE (156.8), an RMSE of 197.5, an $R^2$ of 0.5, and a MAPE of 22.5%.

Across all evaluated models, forecasting accuracy declined as the prediction horizon increased. At shorter horizons, the models produced relatively similar results, particularly for MAE and RMSE, indicating that most algorithms were able to capture short-term changes in boarding time with comparable accuracy. However, as the forecast window extended toward 24 hours, the performance gap between algorithms became more evident. Error metrics, including MAE, RMSE, and MAPE, increased progressively at longer horizons, while $R^2$ values decreased, indicating that the models explained less variation in boarding time as the prediction window became longer. From the 6-hour to the 24-hour horizon, the best MAE increased from 94.2 to 156.8 minutes, the best RMSE increased from 122.6 to 197.5 minutes, MAPE increased from 13.3% to 22.5%, and $R^2$ decreased from 0.8 to 0.5. Linear architectures, particularly NLinear and DLinear, generally maintained lower error values and higher $R^2$ scores than the other evaluated models across the 6-, 8-, 10-, 12-, and 24-hour horizons. NLinear showed the strongest performance at the shorter horizons, while DLinear performed best at the 24-hour horizon. In contrast, TFT, TiDE, and TSTPlus exhibited higher error metrics and lower $R^2$ values overall, with these differences becoming more pronounced at longer forecast horizons. Among these models, TFT showed comparatively higher errors, especially at the 24-hour horizon.

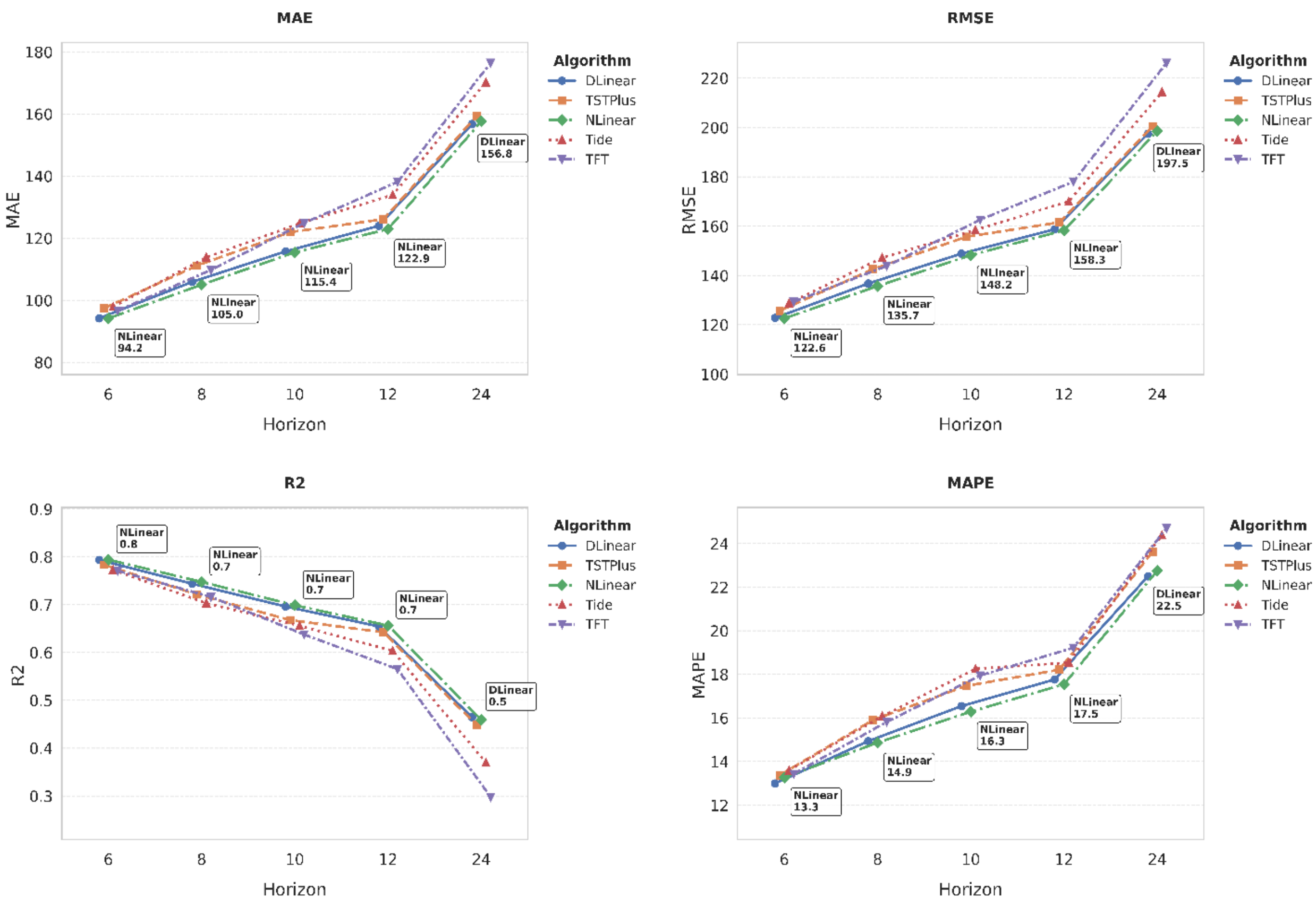

**Fig. 2 |** Multi-horizon forecasting performance for boarding time prediction across five models (DLinear, NLinear, TFT, Tide, and TSTPlus). Results are shown for five forecast horizons (6, 8, 10, 12, and 24 hours) and evaluated using MAE, RMSE, $R^2$, and MAPE.

## Predictive Performance of Extremely Long Boarding Times

Forecasting performance provides an important assessment of model accuracy; however, performance during extreme boarding conditions is of particular operational relevance. Episodes of exceptionally long boarding time reflect periods of heightened strain in the ED and are likely to be the most consequential for proactive decision-making, resource allocation, and escalation planning. Therefore, an additional analysis was conducted to evaluate predictive performance specifically for extreme boarding events, defined using thresholds derived from the observed boarding time distribution.

Extreme cases were identified based on the ground-truth boarding time distribution. Using the observed mean of 622 minutes and standard deviation of 296 minutes, the thresholds were defined as mean plus one standard deviation (622 + 296 = 918 minutes), mean plus two standard deviations (622 + 2 × 296 = 1,214 minutes), and mean plus three standard deviations (622 + 3 × 296 = 1,510 minutes). Hourly observations with boarding time ≥ 918 minutes were classified as Extreme Level 1, ≥ 1,214 minutes as Extreme Level 2, and ≥ 1,510 minutes as Extreme Level 3. In the test dataset, 37.3% of observations met the Level 1 threshold, 7.4% met the Level 2 threshold, and 1.2% met the Level 3 threshold. Table 1 summarizes the best-performing forecasting model and its corresponding MAE at each prediction horizon for Extreme Levels 1, 2, and 3.

Across the defined extreme levels, the best-performing algorithm varied by forecast horizon and severity level. At Extreme Level 1 (mean + SD), NLinear achieved the lowest MAE at the 6-hour, 8-hour, and 12-hour horizons, with MAE values of 102, 114, and 132 minutes, respectively. DLinear performed best at the 10-hour and 24-hour horizons, with MAE values of 124 and 179 minutes, respectively. At Extreme Level 2 (mean + 2SD), NLinear achieved the lowest MAE at 6 hours (144 minutes) and 12 hours (189 minutes), while TiDE performed best at 8 hours (160 minutes). DLinear achieved the lowest MAE at 10 hours (171 minutes) and 24 hours (281 minutes). At Extreme Level 3 (mean + 3SD), DLinear achieved the lowest MAE at 6 hours (187 minutes) and 12 hours (228 minutes), TiDE performed best at 8 hours (182 minutes), and NLinear achieved the lowest MAE at 10 hours (221 minutes) and 24 hours (390 minutes). Overall, linear models, particularly NLinear and DLinear, achieved the lowest errors across most horizon-severity combinations, while MAE generally increased with longer forecast horizons and greater extreme-event severity, reflecting the growing difficulty of accurate prediction under severe boarding conditions.

**Table 1 |** Best-performing model and corresponding MAE for each forecast horizon under increasing extreme boarding thresholds (mean+SD, mean+2SD, and mean+3SD).

| Extreme Level | Algorithm (MAE)* | | | | |
|---|---|---|---|---|---|
| | **6h** | **8h** | **10h** | **12h** | **24h** |
| (1) Mean + SD | NLinear (102) | NLinear (114) | DLinear (124) | NLinear (132) | DLinear (179) |
| (2) Mean + 2*SD | NLinear (144) | Tide (160) | DLinear (171) | NLinear (189) | DLinear (281) |
| (3) Mean + 3*SD | DLinear (187) | Tide (182) | NLinear (221) | DLinear (228) | NLinear (390) |

* All MAE values are reported in minutes.

## Prototype Interface Overview

To assess implementation feasibility and demonstrate how the forecasting framework can be translated into practice, we developed an MLOps-enabled web-based prototype that integrates data flow, forecasting, visualization, experimentation, and retraining within a unified application environment. The application is organized into four modular tabs: Data Stream (Supplementary Fig. S1), Dashboard (Supplementary Fig. S2), Experiment (Supplementary Fig. S3), and Retraining (Supplementary Fig. S4). These components support automated data ingestion, feature generation, forecast delivery, model experimentation, and ongoing model maintenance.

The Data Stream tab (Supplementary Fig. S1) visualizes sequential data ingestion within the forecasting pipeline by simulating the stepwise flow of operational data through the system. It streams patient level ED activity and aggregates hourly operational features to illustrate how raw data are transformed into structured inputs for forecasting. Encounter-level records are displayed as structured cards organized by patient and visit identifiers. Each card summarizes visit-specific operational attributes and timestamps reflecting progression through the ED workflow. This representation illustrates how individual visits are monitored prior to aggregation and modeling. In parallel, the interface presents engineered hourly features that represent the aggregated operational state of the ED. Because these features are computed at an hourly resolution, calculations are performed after the completion of each hourly interval (e.g., data from 9:00 to 10:00 are processed immediately after 10:00) to ensure full information is available for that period. These features serve as direct inputs to the forecasting models and reflect the transformation of raw operational records into structured predictors. Detailed descriptions of the underlying data sources and feature engineering procedures are provided in the Methodology section.

The Dashboard tab (Supplementary Fig. S2) presents boarding time forecasts alongside key operational indicators to support situational awareness and proactive operational review. The interface displays the current aggregated boarding time and corresponding multi-horizon forecasts (6-, 8-, 10-, 12-, and 24-hour ahead predictions), enabling users to compare near-term and extended outlooks within a unified view. In addition to forecast outputs, the dashboard summarizes selected system-level metrics reflecting ED and hospital operational status, including surgical volume, hospital census, waiting count, boarding count, and treatment count. A time-series visualization further contrasts observed boarding time values with forecasted trajectories, facilitating interpretation of short-term predictive trends relative to recent historical patterns. This module illustrates how model outputs can be translated into an interpretable decision-support interface that may help operational users monitor congestion conditions and anticipate the need for escalation planning.

The Experiment tab (Supplementary Fig. S3) provides a configurable environment for dataset construction and model training. This module allows users to configure and evaluate alternative forecasting setups within the application. Feature selection supports inclusion of operational flow metrics and engineered variables, while the target variable can be specified according to the prediction objective. Model training parameters, including lag length, forecast horizon, scaling method, and temporal data range, can be adjusted directly within the interface. Users may also define dataset split proportions for training, validation, and testing to support controlled experimentation. After configuration, the selected forecasting architecture can be trained, and experiment outputs are displayed within the interface to provide immediate feedback on model performance. By enabling model development and refinement within the same environment, this module supports adaptation of the forecasting pipeline as operational needs, data availability, and performance requirements change over time.

The Retraining tab (Supplementary Fig. S4) supports model maintenance after initial development by enabling performance review and controlled model updating as new data become available. This module complements the forecasting and experimentation modules by providing a practical mechanism for sustaining model relevance over time, which is particularly important in dynamic operational environments such as the ED.

Together, the four tabs demonstrate that translating boarding time forecasting into practice requires not only predictive models, but also an integrated application layer that supports data processing, forecast delivery, user interaction, and lifecycle management.

## Discussion

This study advances ED congestion forecasting by systematically benchmarking multiple time series architectures within a unified experimental framework using real-world hospital data. Unlike prior research that focused on descriptive, regression, or simulation-based approaches, we directly evaluated machine learning models for boarding time prediction across multiple forecast horizons. Importantly, the framework relies solely on routinely available operational metrics and publicly accessible contextual data, without using patient level clinical variables, supporting scalability and practical implementation. Beyond predictive benchmarking, this study also addresses a translational challenge that is often underdeveloped in prior work. Making forecasting usable in practice requires not only accurate models, but also an application environment capable of supporting data extraction, forecast delivery, model maintenance, and user interaction.

When interpreted relative to the observed variability in boarding time, model performance remains operationally meaningful. The mean of 622 minutes and a standard deviation of 296 minutes for hourly boarding time indicates substantial temporal dispersion. The best-performing models at each horizon were NLinear at 6, 8, 10, and 12 hours and DLinear at 24 hours. These models achieved MAE values ranging from 94.2 to 156.8 minutes, all of which remained below the observed standard deviation of boarding time. This indicates that, on average, the magnitude of the prediction errors was smaller than the natural variability observed in hourly boarding time. In other words, although ED boarding time showed substantial fluctuation over time, the models were still able to capture meaningful patterns in the data. From an operational perspective, this suggests that the forecasts could provide useful information for anticipating changes in boarding conditions and supporting proactive planning in a highly variable ED environment.

An important implication of these findings relates to threshold-based surge activation strategies. Because boarding time forecasting accuracy declines gradually with increasing horizon length, institutions may consider aligning proactive intervention thresholds within the reliability window of the models. For example, shorter forecast horizons (6–12 hours) may provide sufficient predictive stability to support conditional surge preparation, whereas longer horizons may serve as early situational alerts rather than triggers for immediate escalation. In this way, the forecasting framework may support more proactive operational planning by providing decision makers with advance warning of deteriorating flow conditions before crowding becomes fully manifest. However, translating this capability into routine practice requires the forecasts to be embedded within a usable operational workflow rather than treated as stand-alone model outputs. The extreme case results also underscore the operational value of the proposed framework. Since severe boarding episodes are the most consequential for ED flow management, the ability to maintain useful predictive performance during these periods suggests that the models may help support earlier recognition of deteriorating conditions. This may be particularly relevant for threshold-based surge planning, where forecasts can inform preparatory action before crowding reaches its most critical stage.

Although this study was conducted using data from a single urban, university-affiliated hospital, the proposed framework is designed to support broader institutional applicability. Because the models rely exclusively on routinely available operational metrics and publicly accessible contextual variables, rather than hospital-specific clinical documentation or proprietary data streams, the feature set reflects information commonly available across diverse health systems. This design may enhance portability and facilitate adaptation to institutions with varying electronic health record infrastructures, staffing models, and patient populations. Nevertheless, external validation across multiple hospitals will be necessary before broader generalizability can be assumed.

In addition to evaluating predictive performance, we developed a web-based application prototype to assess the feasibility of operational integration. This prototype serves as a proof-of-concept designed to support preliminary evaluation rather than full deployment. The prototype was intended to provide a unified environment for data streaming, forecast visualization, experimentation, and retraining, reflecting the practical infrastructure needed to translate boarding time forecasting into proactive operational use. As a next step, the system will be presented to clinical and operational stakeholders to obtain structured feedback guided by user-centered design principles. This feedback will inform refinements in system functionality, visualization, and workflow integration prior to broader institutional implementation. In addition, because such systems may influence operational escalation decisions, future implementation should incorporate clear human oversight, ongoing performance monitoring, and governance mechanisms for model updating and use.

Future studies will explore additional features to improve boarding time prediction. While the current models use routinely available operational and system-level metrics, incorporating information such as inpatient discharge timing, unit-level bed availability, elective surgery schedules, and recent congestion patterns may further enhance performance. In addition, the current framework operates at an hourly resolution. Future work will examine finer intervals, such as 15- and 30-minute predictions, to determine whether higher-frequency forecasting improves responsiveness during rapidly changing operational conditions. Future efforts will also extend the current prototype into a more comprehensive decision support system for ED congestion by integrating additional PFMs examined in our previous predictive studies, including boarding count [14] and waiting count [42], while expanding the MLOps framework to support a broader set of congestion-related PFMs.

# Methodology

## Data Source and Feature Engineering

Five data sources were utilized, including internal operational data from a university affiliated urban hospital (ED tracking and inpatient data) and external contextual data from weather services [33], federal holiday calendars [34], and football event schedules [35, 36]. These same sources were used in our prior ED patient flow forecasting studies [14, 37]. All data covered the period from January 1, 2019, to July 1, 2023.

The ED tracking data source captures patient movement and workflow events throughout each visit, with unique patient and visit identifiers enabling chronological tracking from arrival through waiting, treatment, boarding, and discharge. The dataset includes timestamped transitions across ED areas and visit-level attributes such as Emergency Severity Index (ESI) scores (1 = most severe, 5 = least urgent) and reflects operational flow without including clinical information such as laboratory results, imaging findings, or free-text notes. The inpatient unit data source provides hospital-wide operational metrics, including admissions, bed occupancy, discharges, and surgical activity, with timestamps for inpatient arrival, transfers, discharge, and procedure timing when applicable. External contextual variables were incorporated to account for demand drivers outside internal hospital processes, including hourly temperature and categorical weather conditions, federal holidays encoded as binary indicators, and major local Southeastern Conference college football events represented as event indicators.

To enable forecasting, data from all sources were cleaned, aligned, and processed using preprocessing and feature engineering steps consistent with our prior boarding count forecasting work [14]. Feature engineering focused primarily on ED tracking and inpatient unit data. From ED tracking records, we derived hourly measures of waiting, treatment, and boarding processes, including waiting count and time, treatment count and time, boarding count and boarding time, as well as ESI-stratified waiting and boarding counts (ESI 1–2, ESI 3, ESI 4–5). An aggregate ED census feature was computed as the total hourly patient count across waiting, treatment, and boarding states. Count-based variables represent the number of patients present in each state per hour, while time-based variables represent average durations (minutes) calculated by dividing total accumulated time by patient count. Boarding time served as the primary target variable, and an extreme boarding indicator was defined as 1 when hourly boarding time exceeded the mean plus one standard deviation and 0 otherwise. From inpatient unit data, hourly inpatient census and surgical case counts were derived to reflect hospital-wide capacity conditions. External contextual features were engineered at the same hourly resolution, including temperature and categorical weather conditions (clear, clouds, rain, thunderstorm, other), federal holiday indicators (coded as 1 for all 24 hours of recognized holidays), and binary indicators for major local collegiate football events. Finally, temporal features—year, month, day of month, day of week, and hour—were derived from timestamps to capture recurring patterns in ED activity.

After feature engineering, all sources were merged at the hourly level to create a unified forecasting dataset. Preprocessing ensured data quality and operational validity. Visits with waiting times exceeding 9 hours (2.1% of observations) and boarding durations greater than 300 hours were removed as likely anomalies. Data from April to June 2020 were excluded due to atypical conditions during the initial COVID-19 surge (Supplementary Fig. S5). Lagged features were generated using a 24-hour window so that each variable incorporated information from the preceding day. Continuous non-binary features and target variables were standardized using the Standard Scaler [43], with parameters learned from the training set to prevent data leakage, while binary indicators (holiday, football events, weather categories) were left unscaled. The final dataset contained 37,212 hourly observations, with descriptive statistics provided in Figure 1.

Figure 1 summarizes boarding, waiting, treatment, hospital census, and total patient counts including ESI stratified categories, along with associated time-based metrics, temperature, weather categories, and the extreme case indicator. Histograms illustrate empirical distributions, with mean, standard deviation (SD), minimum, and maximum values reported for each feature.

## Training and Evaluation

For training, the lagged features were organized into fixed-length input sequences, with each sample consisting of the previous 24-hourly observations, suitable for supervised time series learning. The resulting tensors followed:

$$X \in R^{N \times d \times 24}, y \in R^{N \times 1} \quad \text{Eq. (1)}$$

where $N$ denotes the number of extracted sliding windows, $d$ the number of features, and 24 the lag length in hours. In our configuration, $d = 30$, yielding input tensors of shape ($N$, 30, 24) for model training. For forecasting, we considered multiple prediction horizons, where the task is to estimate the target value $y_{t+h}$ based on a 24-hour lag window ending at time $t$. Formally, the model learns a mapping with prediction horizons $h \in \{6, 10, 12, 24\}$.

$$\hat{y}_{t+h} = f(x_{t-23:t}) \quad \text{Eq. (2)}$$

Each value of $h$ defines a distinct training and evaluation scenario, enabling assessment of short term and medium-term predictive performance. The dataset was split chronologically into training (70%), validation (15%), and test (15%) sets. Earlier observations were used for model training and hyperparameter tuning, while the most recent data were held out for final performance evaluation.

To benchmark forecasting performance under this formulation, we trained five state of the art deep learning models for multivariate time series forecasting: TiDE, DLinear, NLinear, TFT, and TSTPlus. TiDE employs stacked residual blocks within an encoder decoder architecture that operates on flattened lag windows, using learned projections for multivariate inputs to enable flexible nonlinear mappings from the 24-hour historical context to future ED crowding metrics. DLinear applies explicit series decomposition using a moving average block to separate each input channel into trend and residual components and then learns simple linear mappings over the lag window for each component. NLinear removes the decomposition step and instead applies a single linear layer directly to the standardized lag sequence, optionally centered by the last observation, thereby providing a strong linear baseline. TFT combines recurrent layers, multi-head attention, and gated residual connections with learned variable selection networks to capture both short term dynamics and longer-range temporal dependencies, while retaining interpretability through its attention and gating mechanisms. Finally, TSTPlus is a time series transformer that projects the multivariate input sequence into a learned embedding space with positional encodings, applies stacked self-attention encoder layers, and uses a lightweight pooling head to produce horizon specific forecasts from the same 24-hour input window.

All models were optimized using Optuna [44], with hyperparameters selected separately for each forecast horizon. For TiDE, the hyperparameter search space included learning rate, batch size, dropout, residual block depth, projection widths, and layer normalization. For DLinear, tuning covered learning rate, batch size, moving average kernel size, and design flags related to initialization, weight sharing, and reversible normalization. NLinear tuning focused on learning rate, batch size, and analogous structural configuration flags. For TFT, tuning included learning rate, batch size, dropout, backbone hidden size, number of LSTM layers, number of attention heads, embedding width, and Huber loss parameters with a non-negativity penalty. For TSTPlus, tuning included learning rate, dropout, weight decay, optimizer type, batch size, transformer widths, number of attention heads, number of encoder layers, feed forward width, and activation function. Trials were trained for up to 300 epochs using Adam or related optimizer variants, with early stopping based on validation loss and pruning of underperforming configurations.

Model performances were assessed using standard regression metrics for continuous forecasting: MAE, RMSE, $R^2$, and MAPE. MAE and RMSE capture absolute and squared deviations between forecasts and observed ED metrics, $R^2$quantifies the proportion of explained variability, and MAPE measures proportional error expressed as a percentage. All metrics were computed on the held-out test set for each prediction horizon.

## End-to-End Prototype Architecture

Figure 3 shows the end-to-end system architecture and working prototype developed as part of this project. The system represents a pre-deployment, proof-of-concept implementation designed to simulate hospital operational workflows using historical data. The primary objective is to support stakeholder-facing demonstration and formative evaluation by decision makers at the partner hospital, enabling structured feedback to inform system refinement prior to production deployment.

The prototype consists of three coordinated software services that operate together as an integrated forecasting platform: (1) a data acquisition service responsible for collecting and ingesting raw data; (2) a web application that provides both frontend and backend services for data management, visualization, and inference; and (3) an experiment and retraining service that supports model experimentation, evaluation, and periodic retraining using newly available data. These services were designed to support the practical steps required to translate boarding time forecasting into an operational decision-support environment.

First, a standalone Python-based data acquisition service emulates stepwise data acquisition. It sequentially reads and preprocesses historical data from internal sources, including ED tracking and inpatient systems, as well as external sources such as weather conditions, holidays, and football events. The data acquisition service, shown at the bottom of Fig. 3, transfers the retrieved data into the system, where they are stored as raw records in the corresponding tables of a centralized PostgreSQL database [45].

Second, a web application provides both frontend and backend services. The frontend is organized into four tabs: Dashboard, Data Stream, Experiments, and Retraining supporting visualization, feature monitoring, model experimentation, and retraining, respectively. The backend manages authentication, data access, preprocessing, feature engineering, forecasting, and evaluation. Raw inputs are temporally aligned, aggregated at an hourly resolution, cleaned, normalized, and validated across sources, with engineered features such as waiting count, waiting time, boarding count, and boarding time stored in a centralized PostgreSQL database and exposed through the Data Stream tab. For each prediction horizon ($h \in \{6, 8, 10, 12, 24\}$), a separately trained model generates horizon-specific forecasts, which are saved to the database and displayed in the Dashboard tab alongside observed values and key contextual features at time $t$. Prediction performance is monitored hourly using MAE, $R^2$, and MAPE. When predefined performance thresholds are exceeded, the system is designed to initiate retraining through an asynchronous workflow coordinated via RabbitMQ [46], with additional scheduled monthly retraining to address gradual distributional shifts. This structure was intended to support ongoing model maintenance rather than treating forecasting as a one-time development task.

Overall, the architecture demonstrates that enabling proactive use of boarding time forecasts requires not only predictive models, but also an integrated application layer that supports data processing, forecast generation, visualization, monitoring, and retraining within a single operational workflow. The Experiments tab provides an administrative interface for configuring and executing model development workflows, allowing authorized users to select input features, define target variables, and specify forecasting parameters such as lag window length and prediction horizon. Actions initiated through this interface trigger the experiment and retraining pipeline, which executes dataset construction and model training workflows in the backend.

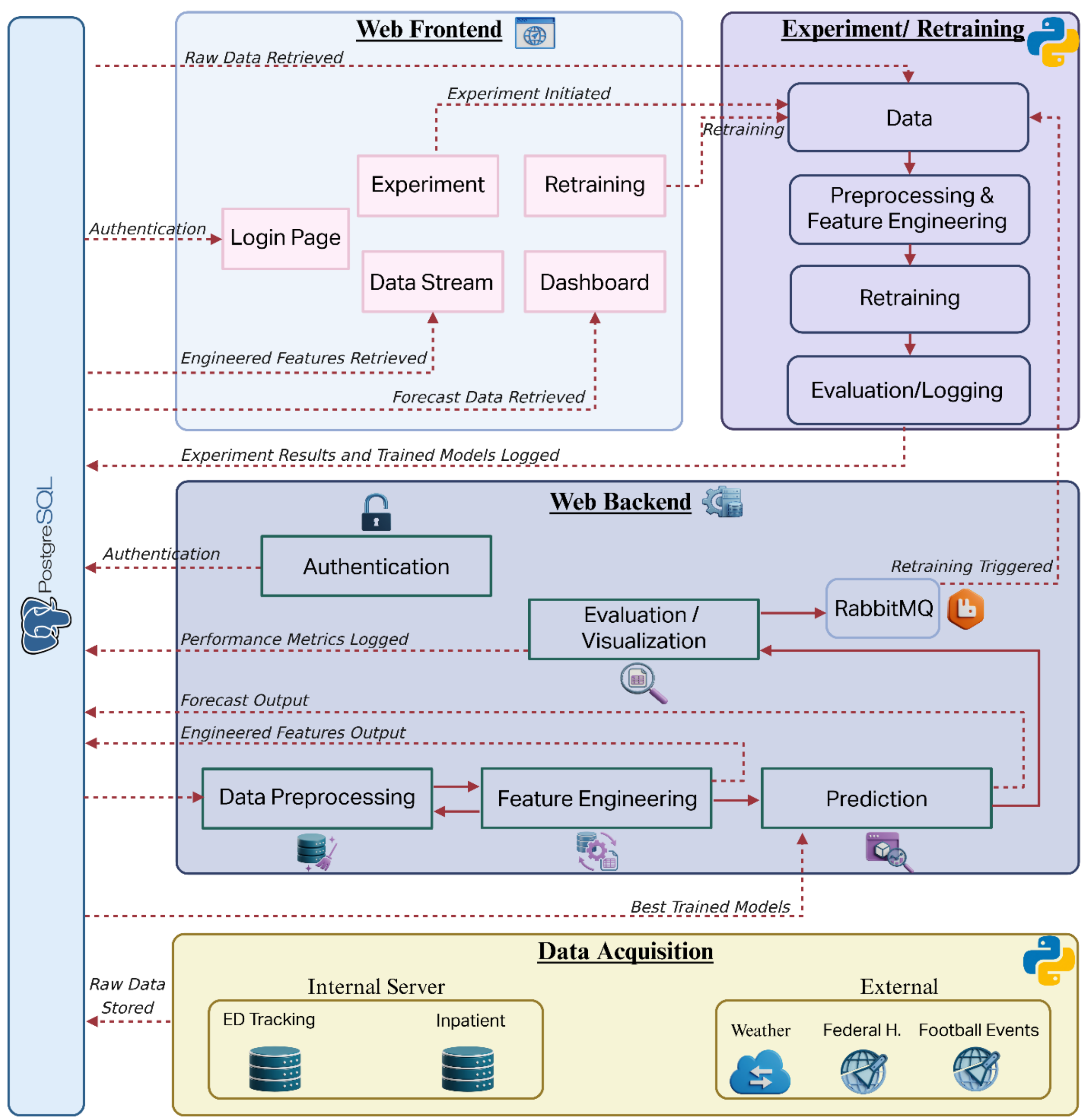

**Fig. 3 |** End-to-end system architecture and proof-of-concept prototype for emergency department boarding time prediction. The platform integrates data acquisition, backend services, a web-based frontend, and model experimentation and retraining pipelines.

## Acknowledgements

This project was supported by the Agency for Healthcare Research and Quality (AHRQ) under grant number 1R21HS029410-01A1.

## Data availability

The data utilized in this study was obtained through a formal collaboration with our partner hospital and is governed by stringent confidentiality agreements, institutional review board (IRB) regulations, and data protection policies. Access to the data is strictly limited to authorized personnel and is intended solely for research purposes under the approved ethical framework.

## Competing interests

The authors declare no competing interests.

## Additional information

Supplementary figures are available at the end of the paper.

## References

[1] A. Watson and W. P. Stuart, "Improving safety and quality with an emergency department overcrowding plan," *Journal of Emergency Nursing,* vol. 49, no. 5, pp. 680-693, 2023.

[2] M. J. Schull, K. Lazier, M. Vermeulen, S. Mawhinney, and L. J. Morrison, "Emergency department contributors to ambulance diversion: a quantitative analysis," *Annals of emergency medicine,* vol. 41, no. 4, pp. 467-476, 2003.

[3] M. Sartini *et al.*, "Overcrowding in emergency department: causes, consequences, and solutions—a narrative review," in *Healthcare*, 2022, vol. 10, no. 9: MDPI, p. 1625.

[4] L. Vanbrabant, K. Braekers, K. Ramaekers, and I. Van Nieuwenhuyse, "Simulation of emergency department operations: A comprehensive review of KPIs and operational improvements," *Computers & Industrial Engineering,* vol. 131, pp. 356-381, 2019.

[5] M. H. Mehrolhassani, A. Behzadi, and E. Asadipour, "Key performance indicators in emergency department simulation: a scoping review," *Scandinavian Journal of Trauma, Resuscitation and Emergency Medicine,* vol. 33, no. 1, p. 15, 2025.

[6] G. Savioli *et al.*, "Emergency department overcrowding: understanding the factors to find corresponding solutions," *Journal of personalized medicine,* vol. 12, no. 2, p. 279, 2022.

[7] C. M. Smalley *et al.*, "The impact of hospital boarding on the emergency department waiting room," *JACEP Open,* vol. 1, no. 5, pp. 1052-1059, 2020.

[8] C. W. Baugh, Y. Freund, P. G. Steg, R. Body, D. J. Maron, and M. Y. A. Yiadom, "Strategies to mitigate emergency department crowding and its impact on cardiovascular patients," *European Heart Journal: Acute Cardiovascular Care,* vol. 12, no. 9, pp. 633-643, 2023.

[9] L. Wu *et al.*, "The association between emergency department length of stay and in-hospital mortality in older patients using machine learning: an observational cohort study," *Journal of Clinical Medicine,* vol. 12, no. 14, p. 4750, 2023.

[10] M. M. Canellas, M. Jewell, J. L. Edwards, D. Olivier, A. H. Jun-O'Connell, and M. A. Reznek, "Measurement of cost of boarding in the emergency department using time-driven activity-based costing," *Annals of Emergency Medicine,* vol. 84, no. 4, pp. 376-385, 2024.

[11] A. Alishahi Tabriz, S. A. Birken, C. M. Shea, B. J. Fried, and P. Viccellio, "What is full capacity protocol, and how is it implemented successfully?," *Implementation Science,* vol. 14, no. 1, p. 73, 2019.

[12] A. L. Nunes, T. Lisboa, B. N. da Rosa, and C. R. Blatt, "Impact of artificial intelligence on hospital admission prediction and flow optimization in health services: a systematic review," *International Journal of Medical Informatics,* vol. 204, p. 106057, 2025.

[13] I. T. Akbasli, A. Z. Birbilen, and O. Teksam, "Artificial intelligence-driven forecasting and shift optimization for pediatric emergency department crowding," *JAMIA open,* vol. 8, no. 2, 2025.

[14] O. Vural, B. Ozaydin, J. Booth, B. F. Lindsey, and A. Ahmed, "Deep Learning-Based Forecasting of Boarding Patient Counts to Address Emergency Department Overcrowding," in *Informatics*, 2025, vol. 12, no. 3: MDPI, p. 95.

[15] M. J. Shaibani, A. A. Fazaeli, R. Daroudi, M. Radinmanesh, and S. Emamgholipour Sefiddashti, "AI-powered models for overcrowding prediction at TUMS hospitals," *Scientific Reports,* 2025.

[16] V. Pasquadibisceglie, A. Appice, D. Malerba, and G. Fiameni, "Leveraging a Large Language Model (LLM) to Predict Hospital Admissions of Emergency Department Patients," *Expert Systems with Applications,* p. 128224, 2025.

[17] E. Benjamin and K. K. Giuliano, "Work systems analysis of emergency nurse patient flow management using the systems engineering initiative for patient safety model: Applying findings from a grounded theory study," *JMIR Human Factors,* vol. 11, no. 1, p. e60176, 2024.

[18] E.-E. Etu *et al.*, "Forecasting Pediatric Emergency Department Arrivals: Evaluating the Role of Exogenous Variables Using Deep Learning Models," *Intelligence-Based Medicine,* p. 100313, 2025.

[19] B. M. Porto and F. S. Fogliatto, "Enhanced forecasting of emergency department patient arrivals using feature engineering approach and machine learning," *BMC Medical Informatics and Decision Making,* vol. 24, no. 1, p. 377, 2024.

[20] H. Hijry and R. Olawoyin, "Predicting patient waiting time in the queue system using deep learning algorithms in the emergency room," *International Journal of Industrial Engineering,* vol. 3, no. 1, pp. 33-45, 2021.

[21] E. L. Williams, D. Huynh, M. Estai, T. Sinha, M. Summerscales, and Y. Kanagasingam, "Predicting Inpatient Admissions From Emergency Department Triage Using Machine Learning: A Systematic Review," *Mayo Clinic Proceedings: Digital Health,* p. 100197, 2025.

[22] L. Poursoltan *et al.*, "Prospective comparison of econometric, machine learning, and foundation models for forecasting emergency department boarding patients," *npj Health Systems,* vol. 2, no. 1, p. 49, 2025.

[23] R. M. Farimani *et al.*, "Models to predict length of stay in the emergency department: a systematic literature review and appraisal," *BMC Emergency Medicine,* vol. 24, no. 1, p. 54, 2024.

[24] W. A. Sulaiman *et al.*, "Leveraging machine learning and rule extraction for enhanced transparency in emergency department length of stay prediction," *Frontiers in Digital Health,* vol. 6, p. 1498939, 2025.

[25] Y.-H. Kuo *et al.*, "An integrated approach of machine learning and systems thinking for waiting time prediction in an emergency department," *International journal of medical informatics,* vol. 139, p. 104143, 2020.

[26] T. Gloyn *et al.*, "Using artificial intelligence to predict patient wait times in the emergency department: A scoping review," *Artificial Intelligence in Medicine,* p. 103316, 2025.

[27] N. Cheng and A. Kuo, "Using Long Short-Term Memory (LSTM) neural networks to predict emergency department wait time," in *The Importance of Health Informatics in Public Health during a Pandemic*: IOS Press, 2020, pp. 199-202.

[28] L. Salehi *et al.*, "Emergency department boarding: a descriptive analysis and measurement of impact on outcomes," *Canadian Journal of Emergency Medicine,* vol. 20, no. 6, pp. 929-937, 2018.

[29] C. K. Prucnal *et al.*, "Patterns in Duration of Emergency Department Boarding and Variation by Sociodemographic Factors," *Western Journal of Emergency Medicine,* vol. 26, no. 6, p. 1640, 2025.

[30] R. Amarasingham, T. Swanson, D. Treichler, S. Amarasingham, and W. Reed, "A rapid admission protocol to reduce emergency department boarding times," *BMJ Quality & Safety,* vol. 19, no. 3, pp. 200-204, 2010.

[31] P. Shi, M. C. Chou, J. G. Dai, D. Ding, and J. Sim, "Models and insights for hospital inpatient operations: Time-dependent ED boarding time," *Management Science,* vol. 62, no. 1, pp. 1-28, 2016.

[32] A. J. Smith *et al.*, "Multisite evaluation of prediction models for emergency department crowding before and during the COVID-19 pandemic," *Journal of the American Medical Informatics Association,* vol. 30, no. 2, pp. 292-300, 2023.

[33] "History Bulk." OpenWeather. https://openweathermap.org/history-bulk (accessed 2025).

[34] "Federal Holidays." United States Office of Personnel Management. https://www.opm.gov/policy-data-oversight/pay-leave/federal-holidays/ (accessed 2025).

[35] "Football Schedule." Alabama Athletics - Official Athletics Website. https://rolltide.com/sports/football/schedule (accessed 2025).

[36] "Football Schedule " Auburn Tigers - Official Athletics Website. https://auburntigers.com/sports/football/schedule (accessed 2025).

[37] O. Vural, B. Ozaydin, K. Y. Aram, J. Booth, B. F. Lindsey, and A. Ahmed, "An Artificial Intelligence–Based Framework for Predicting Emergency Department Overcrowding: Development and Evaluation Study," *JMIR Medical Informatics,* vol. 13, p. e73960, 2025.

[38] A. Das, W. Kong, A. Leach, S. Mathur, R. Sen, and R. Yu, "Long-term forecasting with tide: Time-series dense encoder," *arXiv preprint arXiv:2304.08424,* 2023.

[39] A. Zeng, M. Chen, L. Zhang, and Q. Xu, "Are transformers effective for time series forecasting?," in *Proceedings of the AAAI conference on artificial intelligence*, 2023, vol. 37, no. 9, pp. 11121-11128.

[40] B. Lim, S. Ö. Arık, N. Loeff, and T. Pfister, "Temporal fusion transformers for interpretable multi-horizon time series forecasting," *International journal of forecasting,* vol. 37, no. 4, pp. 1748-1764, 2021.

[41] I. Oguiza. "TSTPlus." Github. https://timeseriesai.github.io/tsai/models.tstplus.html (accessed.

[42] O. Vural, B. Ozaydin, K. Y. Aram, J. Booth, B. F. Lindsey, and A. Ahmed, "An Artificial Intelligence-Based Framework for Predicting Emergency Department Overcrowding: Development and Evaluation Study," *arXiv preprint arXiv:2504.18578,* 2025.
[43] F. Pedregosa *et al.*, "Scikit-learn: Machine learning in Python," *the Journal of machine Learning research,* vol. 12, pp. 2825-2830, 2011.
[44] T. Akiba, S. Sano, T. Yanase, T. Ohta, and M. Koyama, "Optuna: A next-generation hyperparameter optimization framework," in *Proceedings of the 25th ACM SIGKDD international conference on knowledge discovery & data mining*, 2019, pp. 2623-2631.
[45] P. G. D. Group. "PostgreSQL 18.1 Documentation." https://www.postgresql.org/docs/18/index.html (accessed.
[46] "RabbitMQ Documentation." RabbitMQ. https://www.rabbitmq.com/docs (accessed 2026).

# Supplementary

## Figure S1. Data Stream tab of the ED overcrowding system dashboard.

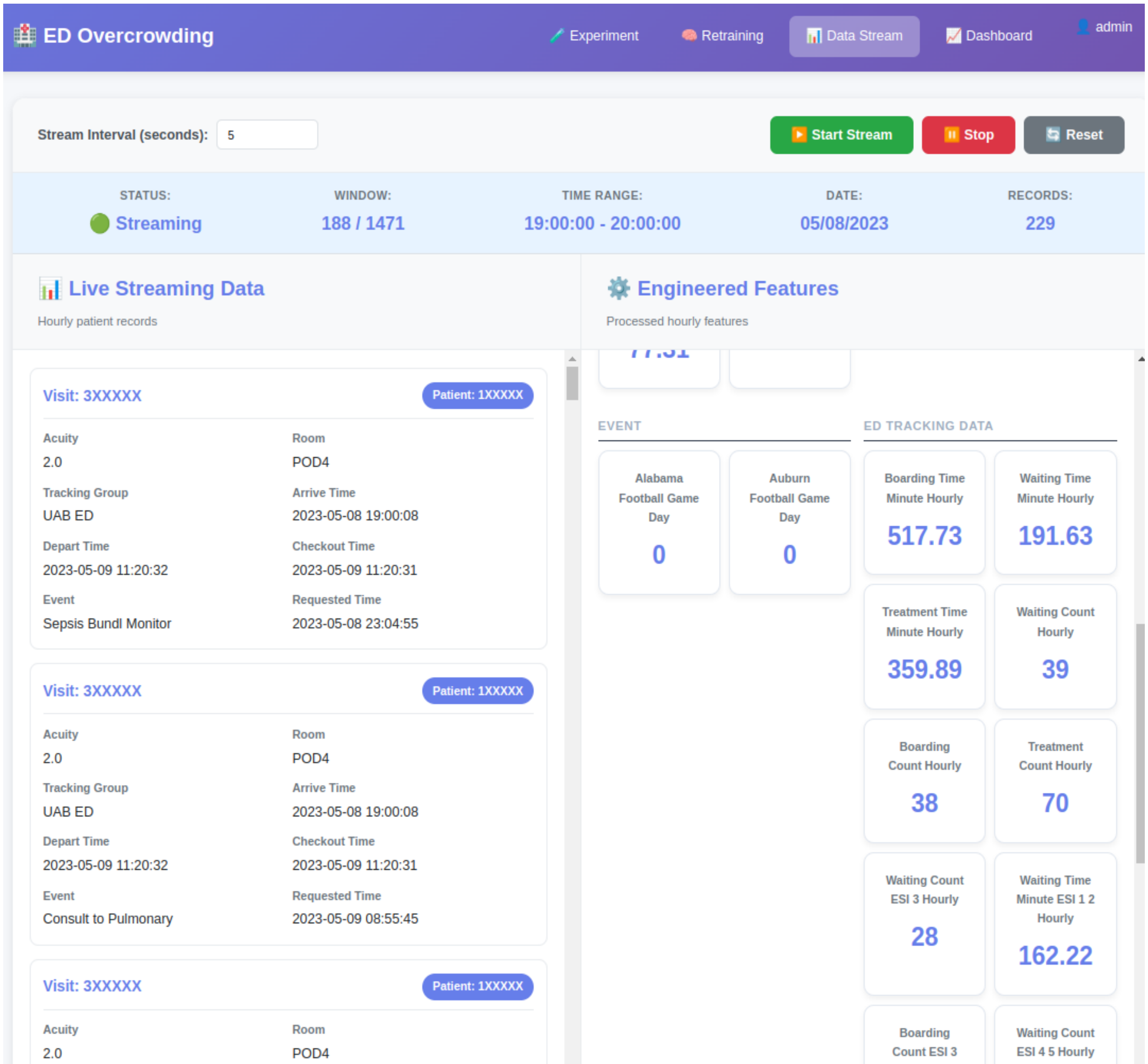

**Figure S2. Dashboard tab of the ED overcrowding system dashboard.**

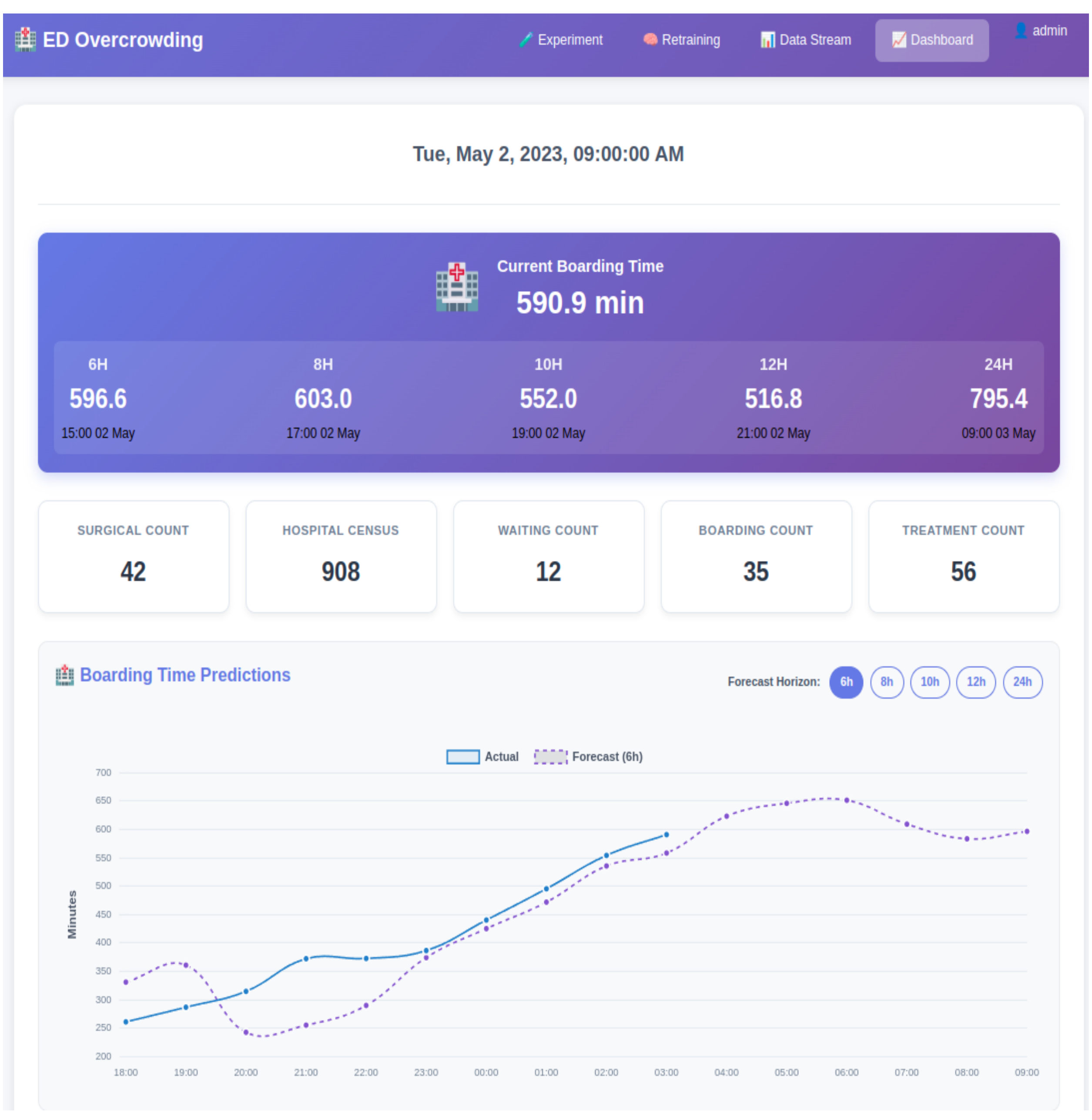

**Figure S3. Dashboard tab of the ED overcrowding system dashboard.**

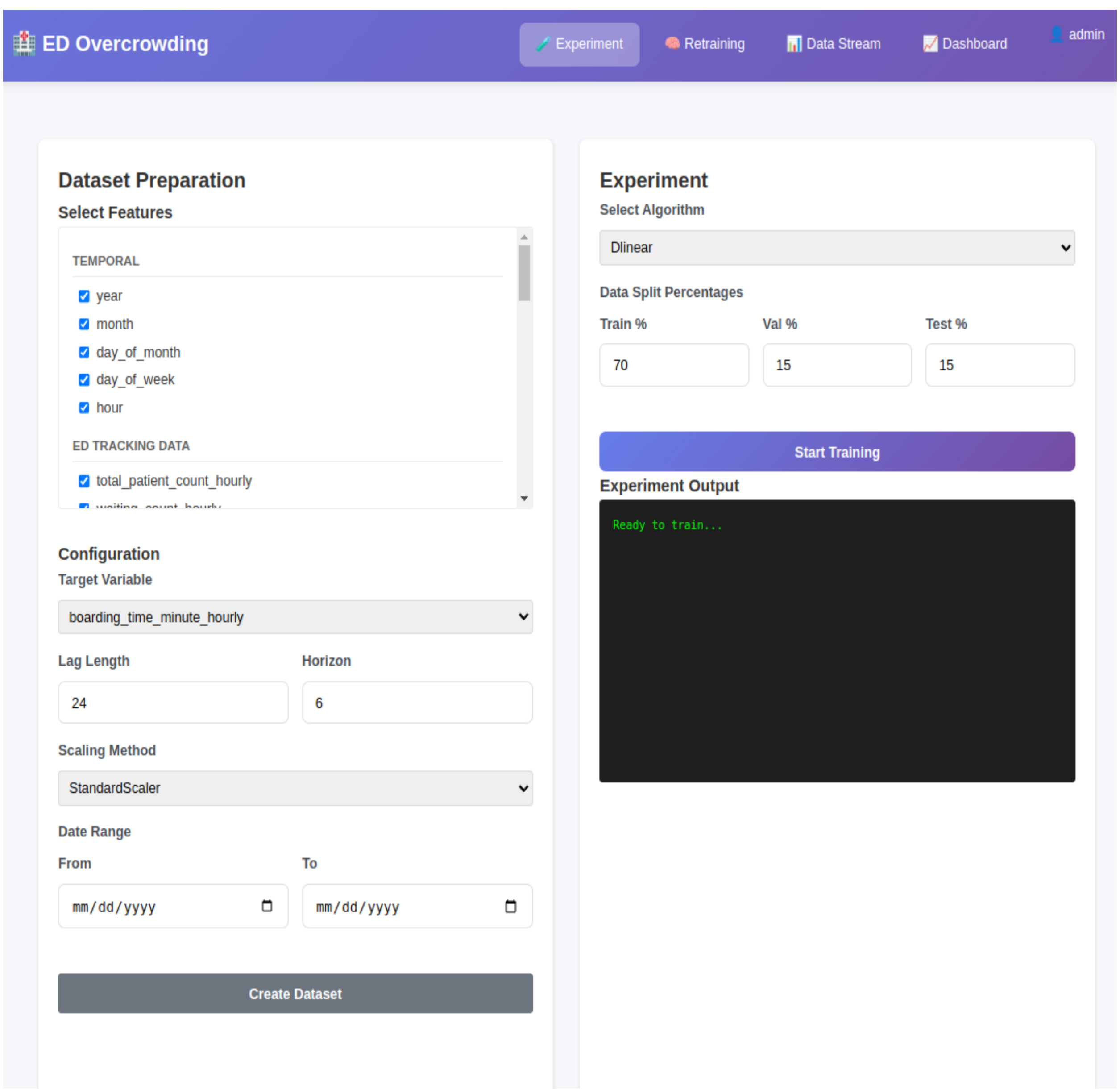

## Figure S4. Retraining tab of the ED overcrowding system dashboard

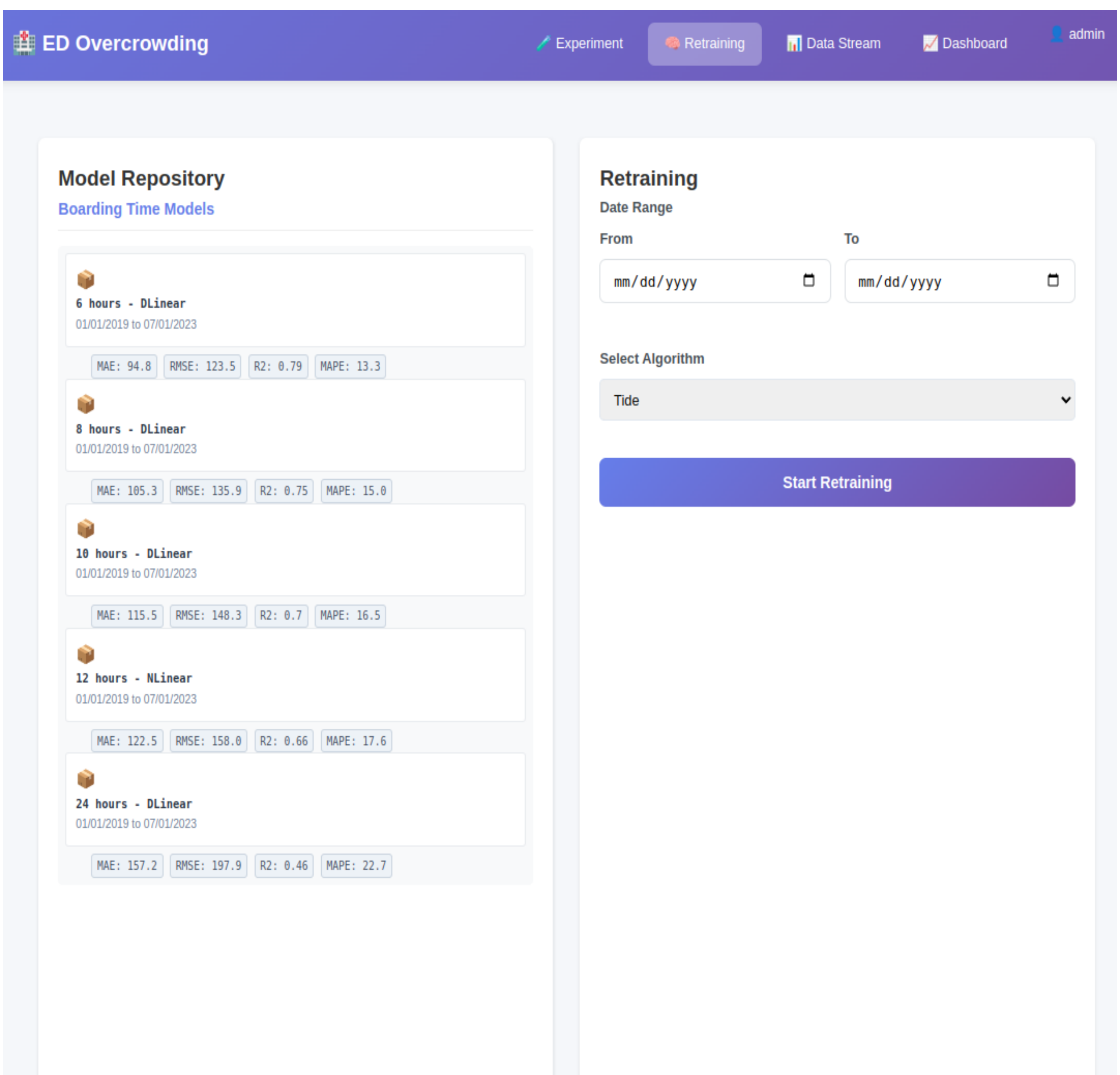

**Figure S5. Boarding Time Trend from January 2019 to June 2023**

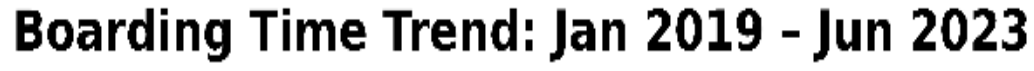

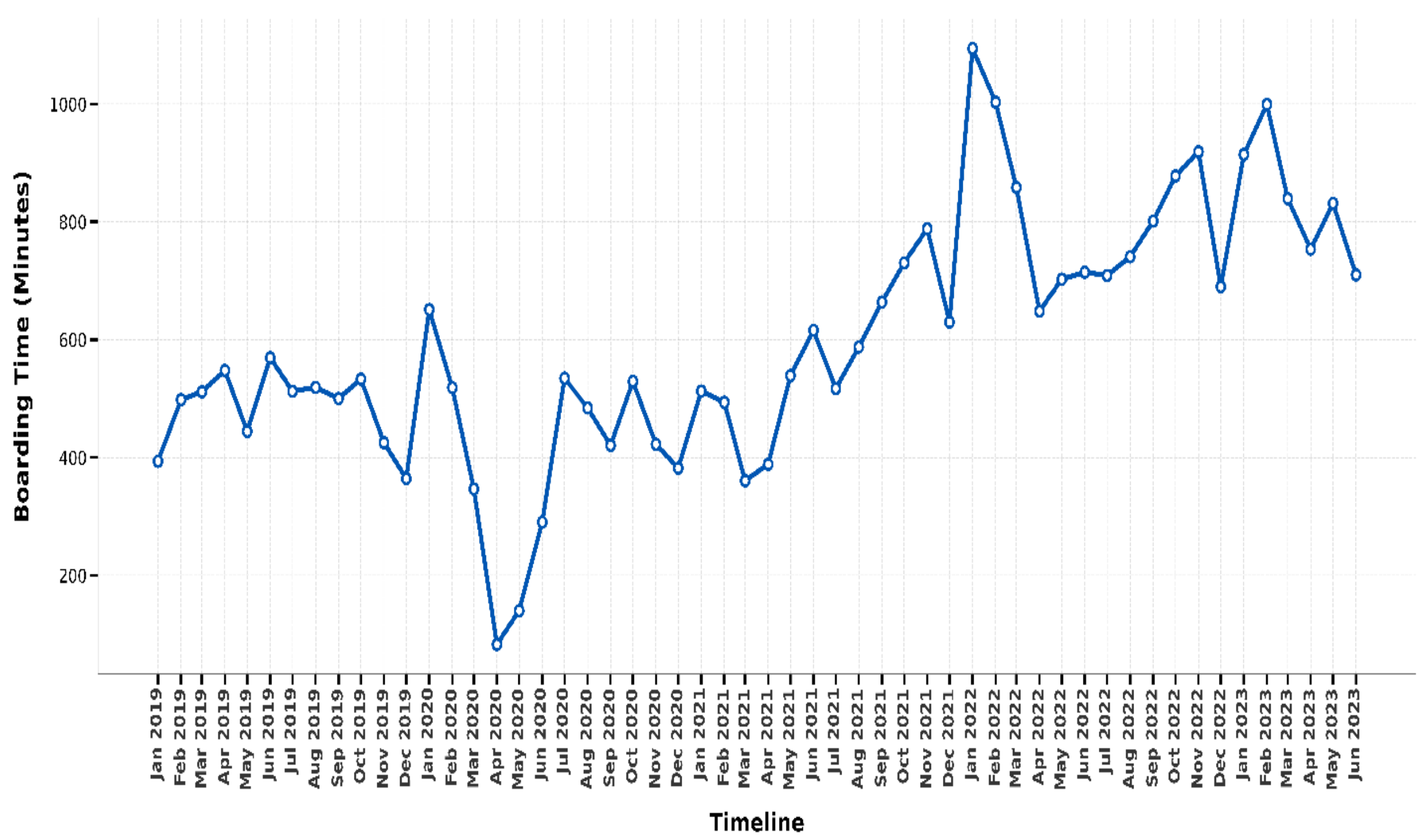